\documentclass{article}

\usepackage{sidecap}
\usepackage{graphicx}
\usepackage{hyperref}

\title{\bf Swarm Systems as a Platform for Open-Ended Evolutionary Dynamics}

\author{Hiroki Sayama$^{1,2}$\\
$^{1}$Binghamton Center of Complex Systems\\
Binghamton University, State University of New York\\
Binghamton, NY 13902, USA\\
$^{2}$Waseda Innovation Lab, Waseda University\\Shinjuku, Tokyo 169-8050, Japan}

\date{}

\begin{document}

\maketitle

\begin{abstract}
Artificial swarm systems have been extensively studied and used in
computer science, robotics, engineering and other technological
fields, primarily as a platform for implementing robust distributed
systems to achieve pre-defined objectives. However, such swarm
systems, especially heterogeneous ones, can also be utilized as an
ideal platform for creating {\em open-ended evolutionary dynamics}
that do not converge toward pre-defined goals but keep exploring
diverse possibilities and generating novel outputs indefinitely. In
this article, we review Swarm Chemistry and its variants as
concrete sample cases to illustrate beneficial characteristics of
heterogeneous swarm systems, including the cardinality leap of design
spaces, multiscale structures/behaviors and their diversity, and
robust self-organization, self-repair and ecological interactions of
emergent patterns, all of which serve as the driving forces for
open-ended evolutionary processes. Applications to science,
engineering, and art/entertainment as well as the directions of
further research are also discussed.
\end{abstract}

\section{Introduction: Swarm Systems and Open-Endedness}

Artificial swarm systems have been extensively studied and used in
computer science, robotics, engineering and other technological
fields, primarily as a platform for implementing robust distributed
systems to achieve pre-defined objectives such as optimization
\cite{ab2015comprehensive}, localization and mapping
\cite{kegeleirs2021swarm}, search and rescue \cite{cardona2019robot}, and
telecommunication network coverage \cite{zhang2020energy}. In these applications, there is always a set of constraints and
objective functions that the designed swarms need to satisfy or
optimize, and convergence to the ``best''
design should be 
achieved as quickly as possible, just like in any other design problems in
engineering and computer science.

In stark contrast to this dominant culture focused on
rapid convergence to high-accuracy solutions, however, we argue in this
article that swarm systems (especially heterogeneous ones) can also
be utilized as an ideal platform for creating {\em open-ended
  evolutionary dynamics} in artificial media. Some researchers
(especially those working in the field of Artificial Life) have
recently been investigating the {\em open-endedness} of artificial
computational systems, namely, how to make such systems {\em not}
converge toward pre-defined goals but keep exploring diverse
possibilities and generating novel outputs indefinitely
\cite{taylor2016open,packard2019overview,pattee2019evolved,sayama2019cardinality,stepney2021modelling,stepney2023open}.
Long-term exploration with no predetermined goals or objectives is exactly
what real biological systems have been performing through evolution
over billions of years, which is the key mechanism for their amazing
adaptability and sustainability \cite{fisher2024sustainability}. The
issue of open-endedness is currently one of the most actively
investigated topics in Artificial Life and AI in general, as it has
potential to create major impacts on future AI systems that are more
creative and more adaptive \cite{stanley2019open,soros2024creativity}.

The possible connection between swarm systems and open-ended evolution was already recognized and investigated by Artificial Life/swarm systems researchers in the late 2000s/early 2010s. Notable early works on this were performed in evolutionary swarm robotics \cite{baele2009open,prieto2010open,bredeche2012environment}, where robotic swarm systems were used as a physical platform on which behavioral algorithms were evolved in an open-ended manner in real time and space without any pre-determined selection criteria. In those earlier works, the number of individual robotic units within a swarm was relatively small. However, as the scale of virtual swarm simulations in Artificial Life research became larger to include hundreds, thousands or even more individuals, researchers came to realize that such large-scale swarms would have great potential to generate and evolve multiscale, emergent, complex dynamical behaviors themselves, and that their creative ability as a collective system was much more substantial than originally imagined \cite{sayama2007decentralized,sayama2009swarm,sayama2011seeking,mototake2015simulation,ikegami2017life,sayama2018seeking,witkowski2019make}. This realization of swarm systems' creative potential grew about the same time as the open-endedness gained its popularity in Artificial Life and AI research communities, which is an interesting coincidence yet naturally bringing these two concepts together.

There are several reasons why large-scale swarms, especially heterogeneous ones,
can be a promising platform for open-endedness. First and foremost,
their design space is vast, which can house a lot of diversity in
structures and dynamic behaviors of self-organizing swarm
patterns. Second, swarm models are scalable in a straightforward way by simply expanding the space
and the size of the swarm population \cite{mototake2015simulation,ikegami2017life,witkowski2019make} and
therefore they can represent multiscale structures/dynamics and their
interactions quite naturally. Third, their self-organizing dynamics
are generally quite robust against perturbations to system
configurations, parameter values or even to spatial dimensionality
\cite{sayama2010robust,sayama2012morphologies}. These key attributes
of swarm systems (vast design space, ability to represent multiscale
structures/dynamics and their interactions, and robustness against
perturbations) are considered to be highly beneficial and useful for realizing open-ended evolutionary exploration in artificial systems \cite{sayama2019cardinality}. Moreover, other benefits of swarm systems include their
clearly visible, understandable mechanisms with all the model parameters
physically interpretable, and their similarity to natural biological
systems and overall familiarity to human users/observers (these are obvious,
given the origins of swarm models \cite{reynolds1987flocks}).

One can evolve swarm systems to maximize or minimize certain well-defined objective metrics
using conventional optimization/evolutionary computation approaches,
but that would not be an adequate approach when swarms are used as a
platform for open-ended evolution. Instead, researchers have adopted
other unconventional approaches, such as connecting swarm evolution
with humans or other exogenous sources of stimuli and removing
explicit fitness evaluation entirely by simulating an evolutionary
ecosystem of swarms as a whole. In particular, the latter approach,
i.e., building Artificial Life systems that evolve according to physically
embodied {\em implicit fitness}, has been demonstrated successfully as
a way to synthesize true {\em natural selection} in artificial media
and thereby autonomously generate various unplanned behaviors through
evolution \cite{ray1991approach,yaeger1994computational,adami1994evolutionary,channon1998evolving}, as the earlier open-ended evolutionary swarm robotics studies also demonstrated successfully \cite{baele2009open,prieto2010open,bredeche2012environment}.

In what follows, we review Swarm Chemistry
\cite{sayama2007decentralized,sayama2009swarm} and its variants as
concrete sample cases to illustrate beneficial characteristics of heterogeneous
swarm systems, including the cardinality leap of design spaces, multiscale
structures/behaviors and their diversity, and robust
self-organization, self-repair and ecological interactions of emergent
patterns, all of which serve as the driving forces for open-ended
evolutionary processes. Applications to science, engineering and
art/entertainment as well as the directions of further research are
also discussed.

\section{Swarm Chemistry and Its Variants}

\subsection{Original Swarm Chemistry}

The original Swarm Chemistry
\cite{sayama2007decentralized,sayama2009swarm} is a heterogeneous
swarm model (largely based on the well-known Boids model
\cite{reynolds1987flocks}) in which multiple distinct types of
self-propelled particles are mixed together to form nontrivial
self-organizing dynamic patterns within a 2D continuous space. The
type of particles that exhibit identical behavior (called ``particle type'' or just ``type'' hereafter) is determined by a combination of unique parameter
values that determine the strengths of individual behaviors such as cohesion,
separation and alignment \cite{reynolds1987flocks}. The specification of a particular
heterogeneous swarm is given in the form of a {\em recipe}, i.e., a list of pairs
of behavioral parameter values and the number of particles that adopt
those parameter values. Such heterogeneous nature of Swarm Chemistry
results in the {\em cardinality leap} \cite{sayama2019cardinality} of the
possibility space, i.e., a fundamental expansion of the cardinality of the possibility sets (such as from countably finite to countably infinite, or from countably infinite to uncountably infinite). This creates the opportunity of open-ended
exploration of various swarm designs.

\begin{figure}[tp]
\centering
\frame{\includegraphics[width=0.161\textwidth]{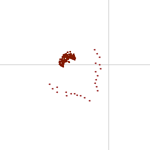}}
\frame{\includegraphics[width=0.161\textwidth]{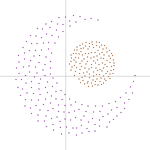}}
\frame{\includegraphics[width=0.161\textwidth]{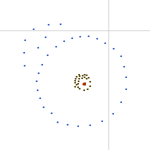}}
\frame{\includegraphics[width=0.161\textwidth]{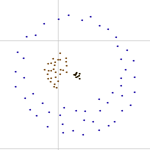}}
\frame{\includegraphics[width=0.161\textwidth]{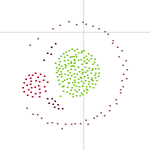}}
\frame{\includegraphics[width=0.161\textwidth]{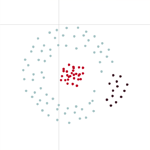}}
\frame{\includegraphics[width=0.161\textwidth]{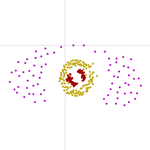}}
\frame{\includegraphics[width=0.161\textwidth]{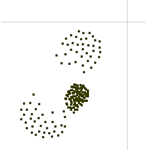}}
\frame{\includegraphics[width=0.161\textwidth]{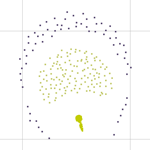}}
\frame{\includegraphics[width=0.161\textwidth]{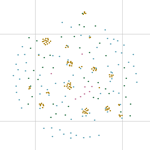}}
\frame{\includegraphics[width=0.161\textwidth]{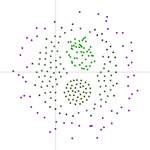}}
\frame{\includegraphics[width=0.161\textwidth]{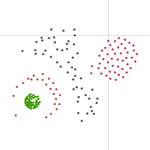}}
\caption{Examples of self-organizing dynamic patterns discovered through
  interactive evolutionary computation in the original Swarm
  Chemistry. Source:
  \url{https://bingdev.binghamton.edu/sayama/SwarmChemistry/\#recipes}. Also
  see some animated videos on
  \url{https://www.youtube.com/@ComplexSystem/search?query=swarm\%20chemistry}.}
\label{original-swarm-chemistry}
\end{figure}

In the original Swarm Chemistry work
\cite{sayama2007decentralized,sayama2009swarm}, such a large design space
of heterogeneous swarms was explored using interactive evolutionary
computation (IEC) \cite{takagi2001interactive,pei2018research}. We developed
both a conventional human-in-the-loop IEC design tool in which users act
as passive fitness evaluators, and a new hyperinteractive evolutionary
computation (HIEC) \cite{bush2011hyperinteractive} design tool in which users play
more active roles in deciding the direction of evolutionary search
rather than being mere fitness evaluators. Using these IEC design tools, a
wide variety of (sometimes highly surprising) self-organizing swarm patterns were
discovered (Fig.\ \ref{original-swarm-chemistry}). These heterogeneous
self-organizing patterns exhibited emergent properties of
macroscopic systems that arose from low-level interactions among
microscopic individual components. They were also found to be
surprisingly robust against the dimensionality change from 2D to 3D
\cite{sayama2012morphologies}, which is quite unique compared to various other complex systems dynamics that heavily depend on the dimensionality of the space (e.g., \cite{montroll1956random,ilachinski2001cellular}). In addition, the interactive swarm
design processes were also utilized as an experimental research tool
to study human collaboration and decision making dynamics
\cite{sayama2009enhancing,sayama2015studying}.

This original Swarm Chemistry model was subsequently modified and
expanded in two separate directions, which are summarized in the
following sections.

\subsection{Morphogenetic Swarm Chemistry}

The first model variant is Morphogenetic Swarm Chemistry
\cite{sayama2010robust,sayama2012swarm,sayama2014four} in which
particles can change their types dynamically (re-differentiation) and
exchange recipes and other observational information within their local
neighbors. Local recipe transmission and stochastic re-differentiation
naturally resulted in swarms' abilities to grow, self-assemble, and even
self-repair after external perturbations were added
\cite{sayama2010robust,sayama2012swarm}.

\begin{figure}[tp]
\centering
\includegraphics[width=\textwidth]{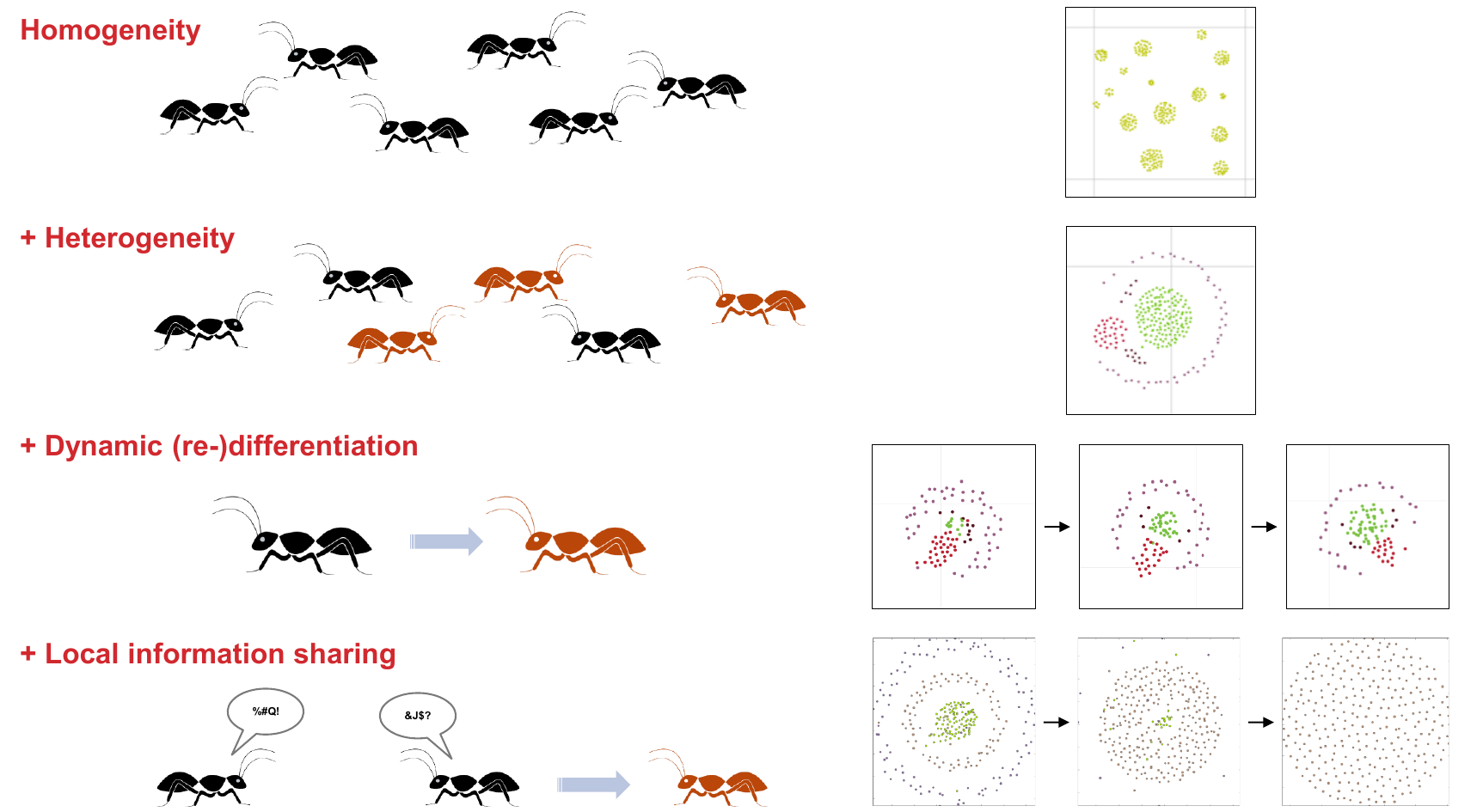}
\caption{Schematic illustrations (left) and snapshots of sample simulation runs (right) of the
  four classes of Morphogenetic Swarm Chemistry systems defined
  according to the individual/collective attributes
  \cite{sayama2014four}. Each attribute builds upon the previous
  one. The most basic swarm is a homogeneous one (1st row). Allowing
  multiple types results in heterogeneous swarms (2nd row), which
  creates room for individuals to dynamically switch types (3rd row),
  which then creates room for them to exchange information locally and
  coordinate their decision making in choosing types (4th row).}
\label{four-classes}
\end{figure}

\begin{SCfigure}[][tp]
\centering
\includegraphics[width=0.5\textwidth]{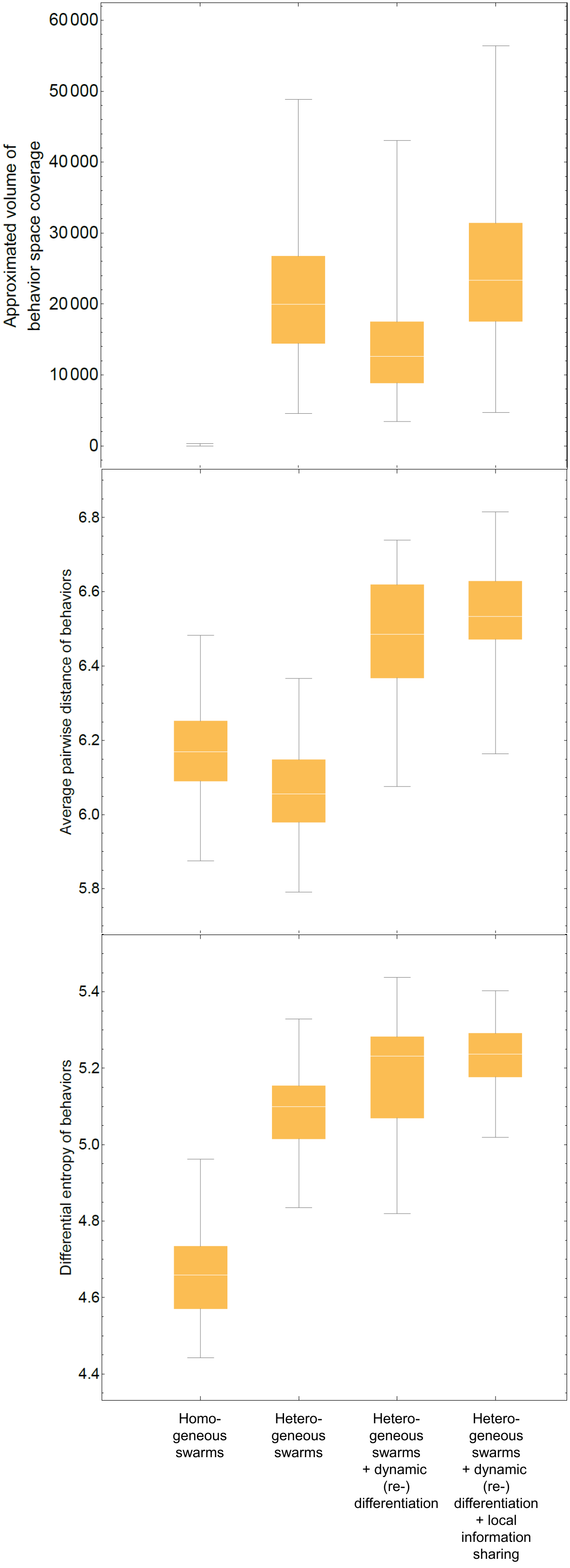}
\caption{Comparison of behavioral diversity measurements across the four classes of Morphogenetic Swarm Chemistry systems illustrated in Fig.\ \ref{four-classes}. Figures were adapted and modified from \cite{sayama2019dynamic}. For each class of swarms, the distributions of measurements were produced by 100 times of bootstrap sampling of 250 out of 500 simulation results. Each simulation was run with a randomly initialized swarm and its behavior was quantified using 24 structural/dynamic characteristics. Top: approximated volume of behavior space coverage. Middle: average pairwise distance of behaviors between randomly selected swarms. Bottom: differential entropy of behavioral distribution. In each plot, greater values indicate greater behavioral diversity of swarms within the class. See \cite{sayama2019dynamic} for more details.}
\label{behavioral-diversities}
\end{SCfigure}

Varying model settings in terms of (1) particle type heterogeneity, (2)
possibility of re-differentiation, and (3) possibility of local
information sharing, allowed for categorization and systematic exploration of four
distinct classes of morphogenetic swarm systems \cite{sayama2014four}
(Fig.\ \ref{four-classes}). Statistical analysis of kinetic and topological characteristics
of self-organizing patterns revealed that the particle type
heterogeneity had major impacts on the structure (topological shape) and behavior (dynamic movement) of the
swarms (e.g., Fig.\ \ref{original-swarm-chemistry}), whereas the re-differentiation and local information sharing abilities had
effects to maintain the swarms spatially more coherent and integrated (e.g., Fig. \ref{four-classes}, 4th row, rightmost image) \cite{sayama2014four}. Moreover, more in-depth analysis of the
generated patterns revealed that the re-differentiation and local
information sharing abilities contributed to the production of more diverse
macroscopic swarm behaviors within each model class (Fig.\ \ref{behavioral-diversities}) \cite{sayama2015behavioral,sayama2019dynamic}. This result indicates
that those behavioral attributes can be helpful in creating and exploring even more
expansive open-ended design spaces than those obtained with the particle type heterogeneity
alone.

\subsection{Evolutionary Swarm Chemistry}

The second model variant is Evolutionary Swarm Chemistry
\cite{sayama2011seeking,sayama2011quantifying}, which was naturally
derived from Morphogenetic Swarm Chemistry described above. The key
model modification was to introduce recipe transmissions (with
possible mutations) between active particles. When
two particles collide, the underlying {\em competition function} decides 
which way the recipe will be transmitted, i.e., which
particle becomes the source (and thus makes the other the destination)
of the recipe transmission, depending on their local
situations. Competition functions are not like conventional fitness
functions but more fundamental ``laws of physics'' that govern the
direction of information flow within the particle population. Several different
competition functions were tested, including the ones that would
choose faster particles, slower particles, particles that hit others
from behind, and particles that were surrounded by more neighbors of
the same type \cite{sayama2011seeking}. Numerical experiments showed
that a version of the latter competition function (called the {\em
  majority} function) was the best in generating most interesting, most
open-ended evolutionary dynamics \cite{sayama2011seeking,sayama2011quantifying}. The majority function chooses the particle that is surrounded by more neighbors of the same type as the competition winner, which tends to promote formation of locally homogeneous but globally heterogeneous patterns with clearly visible macroscopical characteristics. It also makes those patterns more robust against contamination of a small number of external particles with different kinetic parameters. It was also found that high mutation rates and dynamic environmental conditions would further promote continuing evolutionary changes \cite{sayama2011seeking,sayama2011quantifying}.

\begin{figure}[tp]
\centering
\includegraphics[width=0.75\columnwidth]{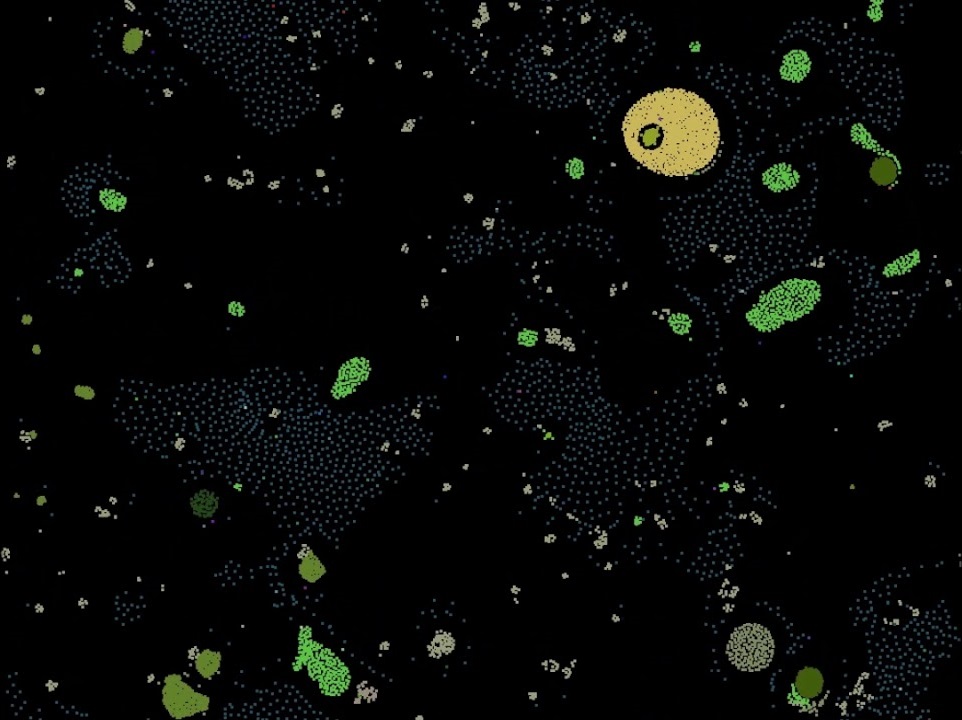}
\caption{A snapshot of a sample Evolutionary Swarm Chemistry
  simulation with 10,000 particles. Colors of particles represent the strengths of their three principal behavioral rules (\{R, G, B\} = \{cohesion, alignment, separation\}), and thus clusters in different colors represent swarms with different behavioral parameters. A wide variety of different swarm
  patterns arise and interact throughout the course of simulation. This
  particular simulation run was initialized using a string ``John Horton Conway'' as
  the seed for random numbers to commemorate his life. Watch the full
  video at \url{https://www.youtube.com/watch?v=YEobkdbCrvQ}.}
\label{swarm-evolution-conway}
\end{figure}

\begin{figure}[tp]
\centering
\includegraphics[width=0.8\columnwidth]{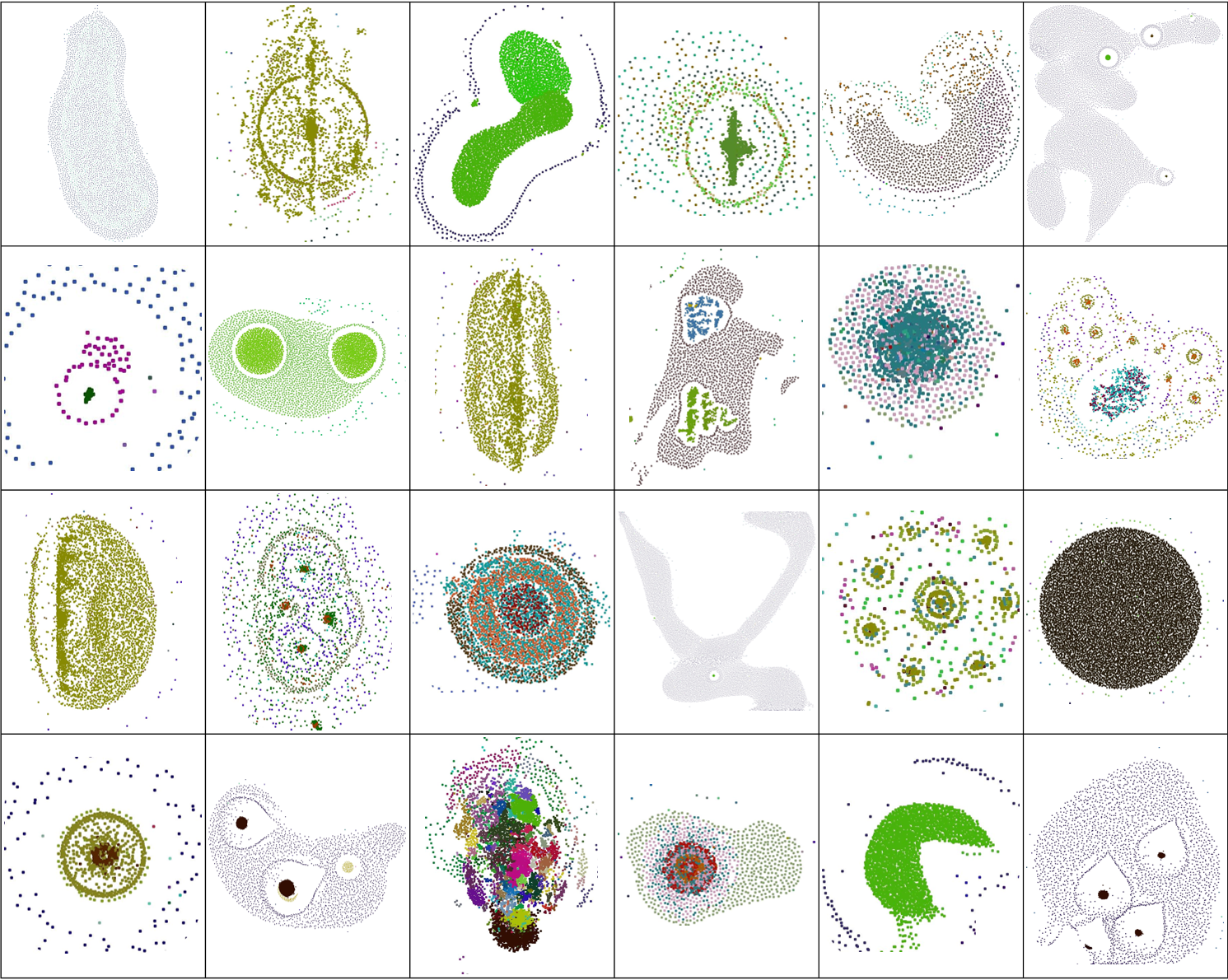}
\caption{Collection of sample swarm patterns that were automatically
  harvested from evolutionary simulation runs
  \cite{sayama2018seeking}, showcasing the inherent creativity of the
  Evolutionary Swarm Chemistry system.}
\label{harvested}
\end{figure}

It is commonly seen in Evolutionary Swarm Chemistry simulations that
visually identifiable ``organisms'' (i.e., swarm patterns made of many particles) chase each
other, consume or get consumed by other organisms, and
self-replicate by binary fissions as they become large. Such emergent behaviors and interactions of higher-level structures produce a surprisingly
rich, complex ecosystem of swarm patterns
(Fig.\ \ref{swarm-evolution-conway}). The ecological and
evolutionary dynamics of macroscopic self-organizing patterns emerge completely
in a bottom-up manner from the lowest level of physics laws governing
information transmission among particles. These evolutionary dynamics
unfold while the total number of particles within the simulated world is
always fixed---no particles will die or be born, and all the evolutionary
dynamics are purely informational and observational. Interestingly, the evolutionary
behaviors of this model would become less manifested in a 3D space
\cite{sayama2012evolutionary}, presumably because the probability for randomly moving particles to encounter would be fundamentally smaller in 3D space than in 2D \cite{montroll1956random}. A more thorough analysis of the
open-endedness of Evolutionary Swarm Chemistry was also conducted using an automated
object harvesting method \cite{sayama2018seeking}, further revealing the
system's inherent creativity (Fig.\ \ref{harvested}). Morphologies of those automatically harvested patterns were further analyzed quantitatively  \cite{sayama2018seeking}, which revealed long-term evolutionary trends of Evolutionary Swarm Chemistry and how its evolutionary outcomes would change dependent on different environmental conditions. For example, under severer environmental conditions, swarms tended to evolve to produce more multicellular-like structures (e.g., rightmost pattern in Fig.\ \ref{harvested}, 2nd row). 

\section{Applications and Expansions}

Since its proposal in the late 2000s, the Swarm Chemistry model has
been used in various scientific and application
research, and the phrase ``swarm chemistry'' (in lower cases) is even becoming a generic term to describe similar kinds of heterogeneous
swarm-based artificial chemistry models in general, not just a pronoun specific to our
models (e.g., \cite{batra2022augmented}). For scientific purposes, Erskine et al.\ \cite{erskine2015cell} used it as a
computational biological model of repeated cell division
dynamics. Nishikawa et al.\ \cite{nishikawa2018exploration} conducted
systematic experiments of the Swarm Chemistry model to investigate how
the asymmetry in kinetic parameters would bring about nontrivial
macroscopic structures and behaviors of a heterogeneous
swarm. Moreover, Rainford et al.\ \cite{rainford2020metachem} used
Swarm Chemistry as one of the benchmarks to evaluate an
algebraic representation framework for artificial chemistry models in
general.

An interesting, very tangible application domain of Swarm Chemistry turns
out to be interactive visual/sound art. Choi and Barger used the
model in multiple interactive art installation/musical live
performance projects
\cite{choi2012playable,choi2014sounds,choi2018interactive}, in which a
human player interacted with the dynamically moving particle swarms to
indirectly control their behaviors and generated sound
patterns \cite{choi2010wayfaring,choi2011wayfaring}. A similar application of Swarm
Chemistry to interactive performance and improvisation was also
performed by Mauceri and Majercik \cite{mauceri2017swarm}.

Other researchers explored different evolutionary search approaches to
find novel swarm patterns that could potentially be used as
distributed control mechanisms of robotic swarms. For example,
Capodieci et al.\ \cite{capodieci2014artificial} used Artificial Immune System methods to evolve Swarm Chemistry
recipes. Nishikawa et al.\ \cite{nishikawa2016coordination} used a
task-oriented optimization approach to evolve coordination mechanisms
of heterogeneous robot swarms in the form of recipes. Similar approaches
of designing and evolving distributed control mechanisms for
physical/engineered heterogeneous robotic swarms have recently
been used to other swarm models in the robotics
community as well
\cite{shirazi2017bio,slavkov2018morphogenesis,mattson2023exploring,vega2023swarm}. In
these applications, the vastness of the design space thanks to the
heterogeneity of components (as originally proposed in Swarm
Chemistry) is both a curse and a blessing that allow for open-ended
exploration and potential innovation to arise.

It is noteworthy that the approaches similar to those developed in the
Swarm Chemistry models are now increasingly adopted in the recent boom
of open-ended distributed dynamical systems in Artificial Life \cite{nichele2023distributed,nichele2024distributed}. For
example, {\em Particle Lenia}
\cite{mordvintsev2022particle,horibe2023exploring}, a swarm-based
variant of the popular continuous cellular automata model {\em Lenia}
\cite{chan2019lenia,chan2020lenia}, produces life-like dynamic
patterns using kinetic interactions among multiple types of particles
in a 2D/3D continuous space. Within the more conventional framework of
cellular automata, {\em Flow Lenia} \cite{plantec2023flow} and {\em
  Evolutionary Lenia} \cite{chan2023towards}, which are recent
extensions of Lenia, aim to create open-ended evolutionary
dynamics of life-like patterns by assigning different parameter values
to specific local cells and letting them propagate through space,
which is similar to the ``recipe'' approach used in Evolutionary Swarm
Chemistry. There are also several other particle-based artificial ecosystem simulations recently created by professional software
developers/hobbyists (not by academic researchers)
\cite{ventrellaClusters,heinemannALIEN,ParticleLifeSimulator,ArtificialLifeSimulator},
which were partly inspired by swarm-based Artificial Life models like
those reviewed in this article. These non-academic projects and
software applications greatly help communicate and popularize the highly attractive, dynamic,
open-ended nature of swarm systems to the general public.

Finally, we also note that very similar self-organizing dynamics of
heterogeneous collective systems have recently attracted substantial
attention in physics, chemistry and material science under
the moniker of ``active matter'' \cite{ramaswamy2017active,liebchen2018synthetic,agudo2019active}. This provides an opportunity to
establish meaningful connections between the Artificial Life/swarm
systems research community and the physics/chemistry/material science research
community, although the extensive body of literature in the
former is still largely unknown to the latter (and vice versa). There are many highly nontrivial, diverse dynamical patterns and behaviors of heterogeneous swarms already discovered in the Artificial Life/swarm systems research, which can be informative and insightful for the active matter community. Meanwhile, the active matter research offers rigorous mathematical frameworks and also has firm grounding on real-world physics/chemistry/biology, from which the Artificial Life/swarm systems research community can learn a lot too. There exists potential
opportunity and need for interdisciplinary discourse and collaboration
among those related fields.

\section{Conclusions: Way Forward}

In this review, we have discussed how swarm systems, especially
those that are heterogeneous, can be useful for designing and
implementing open-ended evolutionary dynamics within computational
systems. We have reviewed Swarm Chemistry and other related systems
as concrete examples to illustrate how such open-ended evolution has
been/can be pursued using swarms. Their massive design space, multiscale
nature, and robustness against perturbations make them one of the most
promising modeling choices for open-endedness research, which is
expected to attract more attention and gain more momentum in the
coming years.

The research reviewed here is still at its nascent stage and there are
many rooms for future work to move the field forward and to make real
societal impacts. The first and most immediate problem is how to make
practical use of emergent patterns arising in open-ended evolutionary
swarms. As indicated in some of the application examples discussed
above, one obvious application domain has been and will continue to be content generation and
interactive entertainment, such as video games, virtual reality, and
computer generated imagery (this was where the original Boids
swarm model \cite{reynolds1987flocks} was also proposed). But are
there other ways to connect open-ended swarm dynamics to more
practical problem-solving applications? What will be ``killer
applications'' of such swarm systems, especially in the era of generative AI?
Answering these questions would bear great importance for the short-
and mid-term success of this research area. 

There can be several potential approaches to address the above ``killer application'' issue, including: (1) developing methodologies to implement custom-made requirements within the open-ended swarm design space so that their emergent dynamics will have greater fit and relevance to a specific real-world problem; (2) integrating mechanistic swarm models with generative AI models so that each model is informed by the other's outputs recursively to create an effective dynamic content generation process; (3) using mechanistic swarm models as a meta-level coordinator of various heterogeneous AI algorithms working in parallel; (4) connecting open-ended swarm systems directly to the real world problem-solving so that their evolution will constantly receive immediate feedback from the reality to drive open-ended exploration and spontaneous solution development; and possibly many others.

The other, more scientifically profound problem is how to establish rigorous,
logical connections between microscopic properties of individuals and
macroscopic emergent properties of collectives. This is a
long-standing ``Big Question'' in complex systems science (often
summarized in Anderson's famous phrase ``More is different''
\cite{anderson1972more}). Heterogeneous swarms' vast design space and
highly nontrivial collective dynamics are, while beneficial for
open-ended evolution, quite a nuisance when one tries to come up with
analytical understanding and explanation of how those complex dynamics
emerge. It will be a challenging yet worthwhile effort to develop a family of mathematical models that can explain emergent structures and dynamics of multi-type heterogeneous swarms (e.g., using partial differential equations of population density functions). However, we may end up finding it nearly impossible to derive
meaningful logical connections between micro and macro for such
complex swarm systems, especially if the number of particle types involved is large and their behavioral interactions are nontrivial. If this turns out to be the case, it will still be helpful
to identify where and how the boundary between explainable and
unexplainable might exist within the spectrum of model
complexity. In either case, the open-ended evolution research can still give us
some hope even if full explanation of swarm systems is not
achievable, because evolution provides ways to create complex systems
even without understanding \cite{sayama2014guiding}.

Finally, we conclude this review by posing yet another problem that
has both practical and scientific importance. That is, we need to
address a fundamental challenge of how to {\em monitor and interpret}
ecological dynamics exhibited by open-ended swarm systems. This might
involve the micro-macro connection problem discussed above, but it
does not have to necessarily; monitoring and interpretation can be
done at macroscopic observational levels only. Even so, however, there
are many open questions on this issue, such as how to identify
``individual organisms'' (e.g., self-organizing local patterns that
behave as a consistent unit) most naturally and how to interpret their
agency (``cognitions'', ``intentions'', ``actions'', etc.) or
reconstruct their evolution (``reproduction'', ``death'',
``variation'', ``genealogy'', etc.) with or without microscopic
details. These properties are often clearly visible to human eyes when
one watches animations of evolutionary simulations of swarms, but
developing objective, algorithmic methods to extract such high-level information
from simulation results would be quite challenging because, ultimately, it is
all purely observational (and this is actually
the case for our real physical/biological worlds too). While some of
the modern AI/machine learning tools like convolutional neural
networks may be helpful for this task, the lack of large-scale
annotated training datasets would be a major roadblock. We envision that developing a
systematic methodology for extracting these higher-level
interpretations from observed swarm dynamics will be quite useful for
both practical applications and scientific investigations, not only
for swarms but also for many other Artificial Life/AI/cognitive
engineering systems and applications.

\section*{Acknowledgments}

The author thanks Robin Bargar, Benjamin James Bush, Insook Choi, Shelley Dionne, Craig Laramee, Howard Pattee, David Sloan Wilson, and Chun Wong for their collaborations and discussions on projects related to swarms and open-ended evolution over many years. This material is based upon work supported in part by the U.S.\ National Science Foundation under Grant No. 0826711 \& 139152, the Binghamton University Evolutionary Studies (EvoS) Small Grant, and the Waseda University Research Promotion Division Research Grant.

\bibliographystyle{plain}
\bibliography{sayama-swarm-review-rev}

\begin{thebibliography}{10}

\bibitem{ab2015comprehensive}
Mohd~Nadhir Ab~Wahab, Samia Nefti-Meziani, and Adham Atyabi.
\newblock A comprehensive review of swarm optimization algorithms.
\newblock {\em PLOS ONE}, 10(5):e0122827, 2015.

\bibitem{adami1994evolutionary}
Chris Adami and C~Titus Brown.
\newblock Evolutionary learning in the {2D} artificial life system {Avida}.
\newblock In {\em Proceedings of Artificial Life IV}. MIT Press, 1994.

\bibitem{agudo2019active}
Jaime Agudo-Canalejo and Ramin Golestanian.
\newblock Active phase separation in mixtures of chemically interacting
  particles.
\newblock {\em Physical Review Letters}, 123(1):018101, 2019.

\bibitem{anderson1972more}
Philip~W Anderson.
\newblock More is different: Broken symmetry and the nature of the hierarchical
  structure of science.
\newblock {\em Science}, 177(4047):393--396, 1972.

\bibitem{baele2009open}
Guy Baele, Nicolas Bredeche, Evert Haasdijk, Steven Maere, Nico Michiels, Yves
  Van~de Peer, Thomas Schmickl, Christopher Schwarzer, and Ronald Thenius.
\newblock Open-ended on-board evolutionary robotics for robot swarms.
\newblock In {\em Proceedings of the 2009 IEEE Congress on Evolutionary
  Computation}, pages 1123--1130. IEEE, 2009.

\bibitem{batra2022augmented}
Sumeet Batra, John Klingner, and Nikolaus Correll.
\newblock Augmented reality for human--swarm interaction in a swarm-robotic
  chemistry simulation.
\newblock {\em Artificial Life and Robotics}, 27(2):407--415, 2022.

\bibitem{bredeche2012environment}
Nicolas Bredeche and Jean-Marc Montanier.
\newblock Environment-driven open-ended evolution with a population of
  autonomous robots.
\newblock In {\em Evolving Physical Systems Workshop}, 2012.

\bibitem{bush2011hyperinteractive}
Benjamin~James Bush and Hiroki Sayama.
\newblock Hyperinteractive evolutionary computation.
\newblock {\em IEEE Transactions on Evolutionary Computation}, 15(3):424--433,
  2011.

\bibitem{capodieci2014artificial}
Nicola Capodieci, Emma Hart, and Giacomo Cabri.
\newblock Artificial immune system driven evolution in {Swarm Chemistry}.
\newblock In {\em Proceedings of the 2014 IEEE Eighth International Conference
  on Self-Adaptive and Self-Organizing Systems}, pages 40--49. IEEE, 2014.

\bibitem{cardona2019robot}
Gustavo~A Cardona and Juan~M Calderon.
\newblock Robot swarm navigation and victim detection using rendezvous
  consensus in search and rescue operations.
\newblock {\em Applied Sciences}, 9(8):1702, 2019.

\bibitem{chan2019lenia}
Bert Wang-Chak Chan.
\newblock Lenia: Biology of artificial life.
\newblock {\em Complex Systems}, 28(3), 2019.

\bibitem{chan2020lenia}
Bert Wang-Chak Chan.
\newblock Lenia and expanded universe.
\newblock In {\em Artificial Life Conference 2020 Proceedings}, pages 221--229.
  MIT Press, 2020.

\bibitem{chan2023towards}
Bert Wang-Chak Chan.
\newblock Towards large-scale simulations of open-ended evolution in continuous
  cellular automata.
\newblock In {\em Proceedings of the Companion Conference on Genetic and
  Evolutionary Computation}, pages 127--130, 2023.

\bibitem{channon1998evolving}
AD~Channon and RI~Damper.
\newblock Evolving novel behaviors via natural selection.
\newblock {\em Proceedings of Artificial Life VI, Los Angeles}, pages 384--388,
  1998.

\bibitem{choi2010wayfaring}
Insook Choi.
\newblock Wayfaring swarms (2010) performance example, 2010.
\newblock \url{https://vimeo.com/25334647}.

\bibitem{choi2011wayfaring}
Insook Choi.
\newblock Wayfaring swarms—concept and technologies, 2011.
\newblock \url{https://vimeo.com/23084185}.

\bibitem{choi2018interactive}
Insook Choi.
\newblock Interactive sonification exploring emergent behavior applying models
  for biological information and listening.
\newblock {\em Frontiers in Neuroscience}, 12:317208, 2018.

\bibitem{choi2012playable}
Insook Choi and Robin Bargar.
\newblock A playable evolutionary interface for performance and social
  engagement.
\newblock In {\em Proceedings of the Intelligent Technologies for Interactive
  Entertainment: 4th International ICST Conference, INTETAIN 2011, Genova,
  Italy, May 25-27, 2011, Revised Selected Papers 4}, pages 170--182. Springer,
  2012.

\bibitem{choi2014sounds}
Insook Choi and Robin Bargar.
\newblock Sounds shadowing agents generating audible features from emergent
  behaviors.
\newblock In {\em Artificial Life Conference 2014 Proceedings}, pages 941--948.
  MIT Press, 2014.

\bibitem{ArtificialLifeSimulator}
{Cikoria Studio}.
\newblock Artificial life simulator.
\newblock
  \url{https://store.steampowered.com/app/2805670/Artificial_Life_Simulator/}.

\bibitem{erskine2015cell}
Adam Erskine and J~Michael Herrmann.
\newblock Cell-division behavior in a heterogeneous swarm environment.
\newblock {\em Artificial Life}, 21(4):481--500, 2015.

\bibitem{fisher2024sustainability}
Len Fisher, Thilo Gross, Helmut Hillebrand, Anders Sandberg, and Hiroki Sayama.
\newblock Sustainability: We need to focus on overall system outcomes rather
  than simplistic targets.
\newblock {\em People and Nature}, 6(2):391--401, 2024.

\bibitem{heinemannALIEN}
Christian Heinemann.
\newblock The alien project.
\newblock \url{https://github.com/chrxh/alien}.

\bibitem{horibe2023exploring}
Kazuya Horibe, Keisuke Suzuki, Takato Horii, and Hiroshi Ishiguro.
\newblock Exploring the adaptive behaviors of {Particle Lenia}: A
  perturbation-response analysis for computational agency.
\newblock In {\em Artificial Life Conference 2023 Proceedings}, volume 2023,
  page~40. MIT Press, 2023.

\bibitem{ikegami2017life}
Takashi Ikegami, Yoh-ichi Mototake, Shintaro Kobori, Mizuki Oka, and Yasuhiro
  Hashimoto.
\newblock Life as an emergent phenomenon: Studies from a large-scale {Boid}
  simulation and web data.
\newblock {\em Philosophical Transactions of the Royal Society A: Mathematical,
  physical and engineering sciences}, 375(2109):20160351, 2017.

\bibitem{ilachinski2001cellular}
Andrew Ilachinski.
\newblock {\em Cellular automata: a discrete universe}.
\newblock World Scientific Publishing Company, 2001.

\bibitem{kegeleirs2021swarm}
Miquel Kegeleirs, Giorgio Grisetti, and Mauro Birattari.
\newblock Swarm {SLAM}: Challenges and perspectives.
\newblock {\em Frontiers in Robotics and AI}, 8:618268, 2021.

\bibitem{liebchen2018synthetic}
Benno Liebchen and Hartmut L\"{o}wen.
\newblock Synthetic chemotaxis and collective behavior in active matter.
\newblock {\em Accounts of Chemical Research}, 51(12):2982--2990, 2018.

\bibitem{mattson2023exploring}
Connor Mattson, Jeremy~C Clark, and Daniel~S Brown.
\newblock Exploring behavior discovery methods for heterogeneous swarms of
  limited-capability robots.
\newblock In {\em Proceedings of the 2023 International Symposium on
  Multi-Robot and Multi-Agent Systems (MRS)}, pages 163--169. IEEE, 2023.

\bibitem{mauceri2017swarm}
Frank Mauceri and Stephen~M Majercik.
\newblock A swarm environment for experimental performance and improvisation.
\newblock In {\em Computational Intelligence in Music, Sound, Art and Design:
  6th International Conference, EvoMUSART 2017, Amsterdam, The Netherlands,
  April 19--21, 2017, Proceedings 6}, pages 190--200. Springer, 2017.

\bibitem{montroll1956random}
Elliot~W Montroll.
\newblock Random walks in multidimensional spaces, especially on periodic
  lattices.
\newblock {\em Journal of the Society for Industrial and Applied Mathematics},
  4(4):241--260, 1956.

\bibitem{mordvintsev2022particle}
Alexander Mordvintsev, Eyvind Niklasson, and Ettore Randazzo.
\newblock Particle {Lenia} and the energy-based formulation, 2022.
\newblock
  \url{https://google-research.github.io/self-organising-systems/particle-lenia/}.

\bibitem{mototake2015simulation}
Yhoichi Mototake and Takashi Ikegami.
\newblock A simulation study of large scale swarms.
\newblock {\em Proc. SWARM}, pages 446--450, 2015.

\bibitem{nichele2023distributed}
Stefano Nichele and et~al.
\newblock The distributed ghost: Cellular automata, distributed dynamical
  systems, and their applications to intelligence (workshop at alife 2023),
  2023.
\newblock \url{https://www.nichele.eu/ALIFE-DistributedGhost/}.

\bibitem{nichele2024distributed}
Stefano Nichele and et~al.
\newblock The distributed viking: Cellular automata, distributed dynamical
  systems, and their applications to intelligence (special session at alife
  2024), 2024.
\newblock \url{https://www.nichele.eu/ALIFE-DistributedViking/}.

\bibitem{nishikawa2016coordination}
Naoki Nishikawa, Reiji Suzuki, and Takaya Arita.
\newblock Coordination control design of heterogeneous swarm robots by means of
  task-oriented optimization.
\newblock {\em Artificial Life and Robotics}, 21:57--68, 2016.

\bibitem{nishikawa2018exploration}
Naoki Nishikawa, Reiji Suzuki, and Takaya Arita.
\newblock Exploration of swarm dynamics emerging from asymmetry.
\newblock {\em Applied Sciences}, 8(5):729, 2018.

\bibitem{ParticleLifeSimulator}
Jairo Oliveira.
\newblock Particle life simulator.
\newblock \url{https://github.com/jairoandre/particle-life}.

\bibitem{packard2019overview}
Norman Packard, Mark~A Bedau, Alastair Channon, Takashi Ikegami, Steen
  Rasmussen, Kenneth~O Stanley, and Tim Taylor.
\newblock An overview of open-ended evolution: Editorial introduction to the
  open-ended evolution {II} special issue.
\newblock {\em Artificial Life}, 25(2):93--103, 2019.

\bibitem{pattee2019evolved}
Howard~H Pattee and Hiroki Sayama.
\newblock Evolved open-endedness, not open-ended evolution.
\newblock {\em Artificial Life}, 25(1):4--8, 2019.

\bibitem{pei2018research}
Yan Pei and Hideyuki Takagi.
\newblock Research progress survey on interactive evolutionary computation.
\newblock {\em Journal of Ambient Intelligence and Humanized Computing}, pages
  1--14, 2018.

\bibitem{plantec2023flow}
Erwan Plantec, Gautier Hamon, Mayalen Etcheverry, Pierre-Yves Oudeyer,
  Cl{\'e}ment Moulin-Frier, and Bert Wang-Chak Chan.
\newblock {Flow-Lenia}: Towards open-ended evolution in cellular automata
  through mass conservation and parameter localization.
\newblock In {\em Artificial Life Conference 2023 Proceedings}, volume 2023,
  page 131. MIT Press, 2023.

\bibitem{prieto2010open}
Abraham Prieto, Jos{\`e}~Antonio Becerra, Francisco Bellas, and Richard~J Duro.
\newblock Open-ended evolution as a means to self-organize heterogeneous
  multi-robot systems in real time.
\newblock {\em Robotics and Autonomous Systems}, 58(12):1282--1291, 2010.

\bibitem{rainford2020metachem}
Penelope~Faulkner Rainford, Angelika Sebald, and Susan Stepney.
\newblock Metachem: An algebraic framework for artificial chemistries.
\newblock {\em Artificial Life}, 26(2):153--195, 2020.

\bibitem{ramaswamy2017active}
Sriram Ramaswamy.
\newblock Active matter.
\newblock {\em Journal of Statistical Mechanics: Theory and Experiment},
  2017(5):054002, 2017.

\bibitem{ray1991approach}
Thomas~S Ray.
\newblock An approach to the synthesis of life.
\newblock In {\em Proceedings of Artificial Life II}, pages 371--408.
  Addison-Wesley, 1991.

\bibitem{reynolds1987flocks}
Craig~W Reynolds.
\newblock Flocks, herds and schools: A distributed behavioral model.
\newblock In {\em Proceedings of the 14th Annual Conference on Computer
  Graphics and Interactive Techniques}, pages 25--34, 1987.

\bibitem{sayama2007decentralized}
Hiroki Sayama.
\newblock Decentralized control and interactive design methods for large-scale
  heterogeneous self-organizing swarms.
\newblock In {\em Proceedings of the 2007 European Conference on Artificial
  Life}, pages 675--684. Springer, 2007.

\bibitem{sayama2009swarm}
Hiroki Sayama.
\newblock {Swarm Chemistry}.
\newblock {\em Artificial Life}, 15(1):105--114, 2009.

\bibitem{sayama2010robust}
Hiroki Sayama.
\newblock Robust morphogenesis of robotic swarms.
\newblock {\em IEEE Computational Intelligence Magazine}, 5(3):43--49, 2010.

\bibitem{sayama2011seeking}
Hiroki Sayama.
\newblock Seeking open-ended evolution in {Swarm Chemistry}.
\newblock In {\em Proceedings of the 2011 IEEE Symposium on Artificial Life},
  pages 186--193. IEEE, 2011.

\bibitem{sayama2012evolutionary}
Hiroki Sayama.
\newblock Evolutionary {Swarm Chemistry} in three-dimensions.
\newblock In {\em Artificial Life Conference 2012 Proceedings}, pages 576--577.
  MIT Press, 2012.

\bibitem{sayama2012morphologies}
Hiroki Sayama.
\newblock Morphologies of self-organizing swarms in {3D Swarm Chemistry}.
\newblock In {\em Proceedings of the 14th Annual Conference on Genetic and
  Evolutionary Computation (GECCO 2012)}, pages 577--584, 2012.

\bibitem{sayama2012swarm}
Hiroki Sayama.
\newblock Swarm-based morphogenetic artificial life.
\newblock In {\em Morphogenetic Engineering: Toward Programmable Complex
  Systems}, pages 191--208. Springer, 2012.

\bibitem{sayama2014four}
Hiroki Sayama.
\newblock Four classes of morphogenetic collective systems.
\newblock In {\em Artificial Life Conference 2014 Proceedings}, pages 320--327.
  MIT Press, 2014.

\bibitem{sayama2014guiding}
Hiroki Sayama.
\newblock Guiding designs of self-organizing swarms: Interactive and automated
  approaches.
\newblock In {\em Guided Self-Organization: Inception}, pages 365--387.
  Springer, 2014.

\bibitem{sayama2015behavioral}
Hiroki Sayama.
\newblock Behavioral diversities of morphogenetic collective systems.
\newblock In {\em Proceedings of the 2015 European Conference on Artificial
  Life}, page~41. MIT Press, 2015.

\bibitem{sayama2018seeking}
Hiroki Sayama.
\newblock Seeking open-ended evolution in {Swarm Chemistry II}: Analyzing
  long-term dynamics via automated object harvesting.
\newblock In {\em Artificial Life Conference 2018 Proceedings}, pages 59--66.
  MIT Press, 2018.

\bibitem{sayama2019cardinality}
Hiroki Sayama.
\newblock Cardinality leap for open-ended evolution: Theoretical consideration
  and demonstration by {Hash Chemistry}.
\newblock {\em Artificial Life}, 25(2):104--116, 2019.

\bibitem{sayama2019dynamic}
Hiroki Sayama.
\newblock Dynamic state transitions of individuals enhance macroscopic
  behavioral diversity of morphogenetic collective systems.
\newblock In {\em From Parallel to Emergent Computing}, pages 105--116. CRC
  Press, 2019.

\bibitem{sayama2009enhancing}
Hiroki Sayama, Shelley Dionne, Craig Laramee, and David~Sloan Wilson.
\newblock Enhancing the architecture of interactive evolutionary design for
  exploring heterogeneous particle swarm dynamics: An in-class experiment.
\newblock In {\em Proceedings of the 2009 IEEE Symposium on Artificial Life},
  pages 85--91. IEEE, 2009.

\bibitem{sayama2015studying}
Hiroki Sayama and Shelley~D Dionne.
\newblock Studying collective human decision making and creativity with
  evolutionary computation.
\newblock {\em Artificial Life}, 21(3):379--393, 2015.

\bibitem{sayama2011quantifying}
Hiroki Sayama and Chun Wong.
\newblock Quantifying evolutionary dynamics of {Swarm Chemistry}.
\newblock In {\em Proceedings of the 2011 European Conference on Artificial
  Life}, pages 729--730. MIT Press, 2011.

\bibitem{shirazi2017bio}
Ataollah~Ramezan Shirazi.
\newblock {\em Bio-inspired self-organizing swarm robotics}.
\newblock University of Surrey (United Kingdom), 2017.

\bibitem{slavkov2018morphogenesis}
Ivica Slavkov, Daniel Carrillo-Zapata, Noemi Carranza, Xavier Diego, Fredrik
  Jansson, Jaap Kaandorp, Sabine Hauert, and James Sharpe.
\newblock Morphogenesis in robot swarms.
\newblock {\em Science Robotics}, 3(25):eaau9178, 2018.

\bibitem{soros2024creativity}
Lisa Soros, Alyssa Adams, Stefano Kalonaris, Olaf Witkowski, and Christian
  Guckelsberger.
\newblock On creativity and open-endedness.
\newblock In {\em Artificial Life Conference 2024 Proceedings}. MIT Press,
  2024.
\newblock In press.

\bibitem{stanley2019open}
Kenneth~O. Stanley.
\newblock Why open-endedness matters.
\newblock {\em Artificial Life}, 25(3):232--235, 2019.

\bibitem{stepney2021modelling}
Susan Stepney.
\newblock Modelling and measuring open-endedness.
\newblock {\em Artificial Life}, 25(1):9, 2021.

\bibitem{stepney2023open}
Susan Stepney and Simon Hickinbotham.
\newblock On the open-endedness of detecting open-endedness.
\newblock {\em Artificial Life}, pages 1--26, 2023.

\bibitem{takagi2001interactive}
Hideyuki Takagi.
\newblock Interactive evolutionary computation: Fusion of the capabilities of
  {EC} optimization and human evaluation.
\newblock {\em Proceedings of the IEEE}, 89(9):1275--1296, 2001.

\bibitem{taylor2016open}
Tim Taylor, Mark Bedau, Alastair Channon, David Ackley, Wolfgang Banzhaf,
  Guillaume Beslon, Emily Dolson, Tom Froese, Simon Hickinbotham, Takashi
  Ikegami, Barry McMullin, Norman Packard, Steen Rasmussen, Nathaniel Virgo,
  Eran Agmon, Edward Clark, Simon McGregor, Charles Ofria, Glen Ropella, Lee
  Spector, Kenneth~O. Stanley, Adam Stanton, Christopher Timperley, Anya
  Vostinar, and Michael Wiser.
\newblock Open-ended evolution: Perspectives from the {OEE} workshop in {York}.
\newblock {\em Artificial Life}, 22(3):408--423, 2016.

\bibitem{vega2023swarm}
Ricardo Vega, Connor Mattson, Daniel~S Brown, and Cameron Nowzari.
\newblock Indirect swarm control: Characterization and analysis of emergent
  swarm behaviors.
\newblock {\em arXiv preprint arXiv:2309.11408}, 2023.

\bibitem{ventrellaClusters}
Jeffrey Ventrella.
\newblock Clusters.
\newblock \url{https://ventrella.com/Clusters/}.

\bibitem{witkowski2019make}
Olaf Witkowski and Takashi Ikegami.
\newblock How to make swarms open-ended? evolving collective intelligence
  through a constricted exploration of adjacent possibles.
\newblock {\em Artificial Life}, 25(2):178--197, 2019.

\bibitem{yaeger1994computational}
Larry~S Yaeger.
\newblock Computational genetics, physiology, metabolism, neural systems,
  learning, vision, and behavior or {PolyWorld}: Life in a new context.
\newblock In {\em Proceedings of Artificial Life III}, pages 263--298.
  Addison-Wesley, 1994.

\bibitem{zhang2020energy}
Xiao Zhang and Lingjie Duan.
\newblock Energy-saving deployment algorithms of {UAV} swarm for sustainable
  wireless coverage.
\newblock {\em IEEE Transactions on Vehicular Technology}, 69(9):10320--10335,
  2020.

\end{thebibliography}

\end{document}